\documentclass[sigconf]{acmart}

\usepackage{booktabs} 
\usepackage{lipsum}
\usepackage{multirow}
\usepackage{array}
\usepackage{courier}
\usepackage{mathtools}
\usepackage{amsmath}
\usepackage{multirow}
\usepackage{mathtools}
\usepackage{amsmath}
\usepackage{float}
\usepackage{graphicx}
\usepackage{algorithm}
\usepackage{algpseudocode}
\usepackage{amssymb}
\usepackage{array}
\usepackage{multirow}
\usepackage{booktabs}
\usepackage{siunitx}
\usepackage{graphicx}
\usepackage{capt-of}
\usepackage{lipsum}

\DeclarePairedDelimiter{\abs}{\lvert}{\rvert}
\setcopyright{rightsretained}

\acmDOI{10.475/123_4}

\acmISBN{123-4567-24-567/08/06}

\acmConference[CIKM'17]{ACM CIKM}{Nov 2017}{Singapore} 
\acmYear{2017}
\copyrightyear{2017}

\acmPrice{15.00}

\begin{document}
\title{Multi-Task Neural Network for Non-discrete Attribute Prediction in Knowledge Graphs}

\author{Yi Tay}
\affiliation{%
  \institution{Nanyang Technological University}
}
\email{ytay017@e.ntu.edu.sg}

\author{Luu Anh Tuan}
\affiliation{%
 \institution{Institute for Infocomm Research}
}
\email{at.luu@i2r.a-star.edu.sg}

\author{Minh C. Phan}
\affiliation{
  \institution{Nanyang Technological University}
}
\email{phan0005@e.ntu.edu.sg}

\author{Siu Cheung Hui}
\affiliation{%
  \institution{Nanyang Technological University}
}
\email{asschui@ntu.edu.sg}

\begin{abstract}
Many popular knowledge graphs such as Freebase, YAGO or DBPedia maintain a list of non-discrete attributes for each entity. Intuitively, these attributes such as height, price or population count are able to richly characterize entities in knowledge graphs. This additional source of information may help to alleviate the inherent sparsity and incompleteness problem that are prevalent in knowledge graphs. Unfortunately, many state-of-the-art relational learning models ignore this information due to the challenging nature of dealing with non-discrete data types in the inherently binary-natured knowledge graphs. In this paper, we propose a novel multi-task neural network approach for both encoding and prediction of non-discrete attribute information in a relational setting. Specifically, we train a neural network for triplet prediction along with a separate network for attribute value regression. Via multi-task learning, we are able to learn representations of entities, relations and attributes that encode information about both tasks. Moreover, such attributes are not only central to many predictive tasks as an information source but also as a prediction target. Therefore, models that are able to encode, incorporate and predict such information in a relational learning context are highly attractive as well. We show that our approach outperforms many state-of-the-art methods for the tasks of relational triplet classification and attribute value prediction. 
\end{abstract}

%




\renewcommand{\footnotesize}{\scriptsize}





\maketitle

\section{Introduction}
Knowledge graphs (KGs) are semantic networks that are highly popular in many intelligent applications such as question answering, entity linking or recommendation systems. KGs are typically expressed as triplets in the form of (head, relation, tail) where each entity may be connected to another entity via a semantic relation, e.g., (SteveJobs, \textit{isFounderOf}, Apple). Knowledge graphs, aside from being a popular universal representation of knowledge such as ConceptNet or WordNet, are also widely adopted in practical applications. Examples of popular large scale knowledge graphs include YAGO, Freebase and Google Knowledge Graph. Given the immense popularity in many AI applications, reasoning with knowledge graphs is an extremely desirable technique. In this context, machine learning algorithms for relational reasoning are highly sought after. 

At its core, relational learning aims to reason and infer probabilistic estimates regarding new or unseen facts in knowledge graphs. Intuitively, this supports applications ranging from knowledge graph completion to recommendation systems. Recently, many state-of-the-art relational learning techniques \cite{DBLP:conf/nips/BordesUGWY13,DBLP:conf/aaai/LinLSLZ15,DBLP:conf/nips/SocherCMN13,DBLP:conf/kdd/0001GHHLMSSZ14} have been proposed. These techniques, which can also be considered as representation learning techniques, are concerned with modeling multi-relational data in latent space and learning continuous vector representations of entities and relations in KGs. Through reasoning in latent embedding space, probabilistic estimates can be inferred which contribute to alleviating the intrinsic problem of knowledge graph incompleteness. Furthermore, relations in knowledge graphs can be extremely sparse \cite{DBLP:journals/corr/XieMDH17} in which relational information might be insufficient for making solid predictions or recommendations.  

Unfortunately and additionally, many of the state-of-the-art approaches are solely focused on exploiting structural information and \textbf{neglect} the rich attribute information that are readily available in knowledge graphs. Intuitively, attribute information can richly characterize entities that would help combat relational sparsity. As such, in this paper, we propose that this information should be and can be easily incorporated into relational learning algorithms in order to improve the relational learning and at the same time, enable prediction of these non-discrete attribute information. To this end, we propose an end-to-end multi-task neural network. To give readers a better insight behind our intuitions and motivations behind this work, we begin by providing some context pertaining to the problem at hand.

\subsection{Motivation}
 Firstly and naturally, non-discrete attributes are commonplace in many KGs. For example, non-discrete attributes such as the price of a product (\textit{iPhone}, hasPrice, $1000$) or the height of an actor (\textit{DanielRadcliffe}, hasHeight, \textit{1.7m}) are often present in knowledge graphs. As such, our work is mainly concerned with both leveraging these non-discrete attributes to improve the relational learning process and enable prediction of these attributes. To this end, we answer the following four important questions:
\begin{enumerate}
\item Can attribute information really help in relational learning?
\item Why do we want to predict attributes? 
\item Are there so many attributes in KGs? 
\item Why is this difficult?
\end{enumerate}

\subsubsection{Can attribute information really help in relational learning?}
 Consider a relational learning task of completing the following triplet - (person, \textit{hasGender}, ?). Traditionally, relational learning exploits statistical patterns across relations to make predictions. The rule \textbf{R1:} (personA, \textit{isMotherOf}, personB) $\rightarrow$ (personA, \textit{hasGender}, Female) is an example of what relational learning models are capable of learning in order to make predictions. However, statistical information available through relations can be really sparse. Additionally, knowledge graphs can also be highly incomplete. As such, there is a good chance that, at test time, an entity arrives without any relation such as \textit{(isMotherOf)} to make an informed decision. This is where attribute information can be complementary and also enhance the relational learning model. In this case, it might be possible that the \textit{(hasHeight)} attribute alone is able to allow the model to predict the gender correctly. Intuitively, attribute information can richly characterize entities in a knowledge graph. In the studied example, the non-discrete value of the attribute \textit{(hasHeight)} is strongly correlated with the relation \textit{(hasGender)}. In another context, the height or weight attributes of entities (such as products) may allow relational learning models to differentiate between cars and iphones. Intuitively, this can be also seen as leveraging entity schema information. Therefore, we believe that attribute information that can richly characterize entities will provide relational learning models with valuable knowledge. 

\subsubsection{Why do we want to predict attributes?}
Additionally, the importance of non-discrete attributes goes beyond serving as an auxiliary information source but also as a prediction target. There are mainly two ways to interpret this task. Firstly, it can be interpreted as using relational structural information as features for a regression task which forms a huge chunk of standard machine learning tasks in both research and industry. In this case there is a target attribute that is important to be predicted such as sales forecast, the GPA of a student, or prices of products. Secondly, the motivation of attribute prediction can also be considered identical to that of knowledge graph completion for relational triplets, e.g., a person's height might be missing in the knowledge graph and attribute value prediction can help infer this information. 

\subsubsection{Are there so many attributes in KGs?}
For each \textbf{entity} in knowledge graphs such as Freebase, a list of relevant attributes are maintained. In the Freebase dump \textit{Easy Freebase} \cite{DBLP:conf/www/BastBBH14}, there are already $27$ million non-discrete attribute triplets ($\approx 10\%)$. In our extracted Freebase subgraph, we easily obtain a percentage of $\approx 33\%$ non-discrete attribute triplets. Hence, this further motivates our problem, i.e., we are ensured that there is sufficient and reasonable amount of attribute information. Additionally, the non-discrete nature of attributes suggests that each triplet contains more encoded information over binary-natured relational triplets. Table \ref{tab:example_attr} describes some common non-discrete attributes and their respective entity types from Freebase. 

\begin{table}[htbp]
  \centering
  \small
    \begin{tabular}{ccc}
    \hline
    Attribute & Value  & Entity Type\\
    \hline
         num\_employees & 1204 & Company\\
          num\_postgraduates & 120 & University\\
          length & 10 & Movies (Duration)\\
          career\_losses\_single & 50 &Tennis players\\
          age & 52 & People\\
          height & 170 & People\\
          latitude & 1.783874 & Location \\
          \hline
    \end{tabular}%

  \caption{Sample non-discrete attributes and their values on Freebase and YAGO.}
    \label{tab:example_attr}%
\end{table}%

\subsubsection{Why is this difficult?}
Knowledge graphs are typically considered as 3D graphs, i.e., (head, relation, tail). Non-discrete attributes, can often act as relations in which their attributes are binned and casted as entities. However, this can easily cause the size of the knowledge graph to blow up and aggravate the scalability issue faced by many relational learning techniques. Attributes, intuitively speaking, seem to cast knowledge graphs out of its comfort zone of binary truths (0 or 1). As such, there are limited works that attempt to exploit attribute information in the setting of KGs \cite{DBLP:conf/www/NickelTK12}. While this has been done on bipartite graphs or similar networks with only one entity type, the diversity of entity types in KGs can be large. In these cases, the feature vectors of attributes can be extremely sparse and ineffective when there are a significant number of entity types in the knowledge base. Moreover, incorporating feature vectors of attributes will not enable the model to predict attribute values. Overall, it is evident that many of the current approaches do not handle such information and more often than not discard these attributes completely during the training process. 

\subsection{Contributions}
In this paper, we propose a neural network approach that elegantly incorporates attribute information and enables regression in the context of relational learning. The primary contributions of this paper are as follows: 

\begin{itemize}
\item We propose a novel deep learning architecture for representation learning of entities, relations and attributes in KGs. Our unified framework which we call \textit{Multi-Task Knowledge Graph Neural Network} (MT-KGNN) trains two networks in a multi-task fashion. Via using a shared embedding space, we are able to encode attribute value information while learning representations. Furthermore, our network also supports the prediction of attribute values which is a feature lacking in many relational learning methods. 
\item For the first time, we evaluate the ability of many relational learning approaches such as \cite{DBLP:conf/nips/BordesUGWY13,DBLP:conf/icml/NickelTK11} in the task of attribute value prediction. This is a new experiment designed to evaluate the competency and suitability of KG representation learning techniques in encoding attribute information. 
\item Aside from being able to perform regression typed predictions in a relational context, our approach also demonstrates state-of-the-art performance on traditional relational learning benchmarks such as triplet classification.
\end{itemize}

\section{Related Work}
Across the rich history of relational learning, there have been a myriad of machine learning techniques proposed to model multi-relational data. While traditional relational learning dates back to probabilistic, feature engineering or rule mining approaches, it has been recently fashionable to reason over knowledge graphs in latent embedding space. This field, also known as deep learning or representation learning, is responsible for many state-of-the-art applications in the fields of NLP and Artificial Intelligence.

The popular latent embedding models can be generally classified yet again into three categories, namely the \textit{neural network} approach, the \textit{knowledge graph embedding} approach and finally the \textit{factorization} approach. Before we briefly describe each category, let us begin with a formal definition of the notation that will be used throughout this paper.

\subsection{Problem Formulation and Notation}
In this section, we briefly describe the notation used in this paper along with a simple introduction of relational learning problems. 
\subsubsection{Notation}
Let $\Delta=(E,R)$ denote a knowledge graph. $E=\{e_1,e_2...e_{\abs{E}}\}$ is the set of all entities, and $R=\{r_1,r_2...r_{\abs{R}}\}$ is the set of all relations. A relational triplet $\xi_i \in \xi$ is defined as $(e_i, r_k, e_j)$ where two entities are in a relation $r_k$ and can be interpreted as a fact that exists in the knowledge graph $\Delta$. Since attributes are specific to our model, we introduce them in later sections. For latent embedding models, entities $e_i \in E$ and relation $r_i \in R$ are often represented as real-valued vector or matrix representations. In order to facilitate such a design, embedding matrices $W_e \in \mathbb{R}^{\abs{E} \times n}$ and likewise, $W_r \in \mathbb{R}^{\abs{R} \times n}$ are parameters of the model. To select the relevant entity or relation, we can simply perform a look-up operation via one-hot vector. As such, the inputs to our model are often represented as indices that map into these embedding matrices.
\subsubsection{Problem Formulation}
There are mainly two popular relational learning tasks, namely triplet classification and link prediction. Simply speaking, the goals of both tasks are similar and are complementary to many semantic applications. Making probabilistic recommendations and mitigating the inherent incompleteness of knowledge graphs are examples of such applications. As such, the crux of many relational learning tasks is to produce a score
\begin{equation}
 s(e_i, r_k, e_j) \: \in \: [0,1]
 \end{equation}
 that denotes the probability or strength of a fact or triplet. Naturally, similar to many ranking problems or classification problems, this task can be casted into three different variations, i.e., pointwise, pairwise and listwise. Pointwise considers a simple binary classification problem in which a sigmoid or 2-class softmax layer might be used. Pairwise tries to maximize the margin between the golden facts and negative (corrupted) samples. The pairwise hinge loss can be described as follows: 
\begin{equation}
L =  \sum_{\xi \in \Delta} \sum_{{\xi}' \not\in \Delta} max(0,E_l(\xi) + \lambda - E_l({\xi}'))
\end{equation}
Finally, listwise considers all candidate entities, i.e., for $(e_i, r_k, ?)$, consider all $e_j \in E$. Since the listwise approach is rarely used in the literature of relational learning, we do not consider the listwise approach in our paper. Moreover, the pairwise and listwise approaches require a significant amount of implementation effort over the pointwise approach and as such, we consider a pointwise approach when designing our model. Finally and also intuitively, the pairwise and listwise approaches are also not suitable for attribute value regression.

\subsection{Neural Networks for Relational Learning}
In this section, we give a brief overview of the existing state-of-the-art neural network (NN) approaches for relational learning. Neural networks are connectionist models that form the heart of the deep learning revolution. The key driving force behind these models are that the parameters are learned via stochastic gradient descent (SGD) and backpropagation. There are two popular neural network approaches for relational learning. The first approach is the ER-MLP (\textit{Entity Relation Multi-Layered Perceptron}) which was incepted as part of the Google Knowledge Vault Project \cite{DBLP:conf/kdd/0001GHHLMSSZ14}. The other approach is the Neural Tensor Network (NTN) \cite{DBLP:conf/nips/SocherCMN13} which models multiple views of dyadic interactions between entities with tensor products. 

\subsubsection{ER-MLP}
ER-MLP was studied as a simple neural baseline for relational learning in the Google Knowledge Vault Project. The scoring function of ER-MLP can be represented as follows:
\begin{equation}
s(e_i,r_k,e_j) = \sigma(\vec{v}^{\top}f(\textbf{W}_{\phi}^{\top}[\vec{e_i};\vec{e_j};\vec{r_k}]))
\end{equation}
where $f$ is a non-linear activation function such as $tanh$ applied element-wise. $v \in \mathbb{R}^{h \times 1}$ and $\textbf{W}_{\phi} \in \mathbb{R}^{3n \times h}$ are parameters of the network. $\sigma$ is the sigmoid function. $h$ is the size of the hidden layer and $n$ is the size of the embeddings. $[;]$ denotes a concatenation operator. Intuitively, ER-MLP extracts the embeddings of the entities and relations involved in the triplet and simply concatenates them. Despite its simplicity, it has been proven to be able to produce highly competitive performance in relational learning tasks. 

\subsubsection{Neural Tensor Network (NTN)}
This highly expressive neural model was first incepted in the fields of NLP for sentiment analysis and subsequently applied in multiple domains such as knowledge base completion \cite{DBLP:conf/nips/SocherCMN13} and question answering \cite{DBLP:conf/ijcai/QiuH15,DBLP:conf/sigir/TayPLH17}. The key intuition behind NTN is that it models multiple views of dyadic interactions between entities with a relation-specific tensor \cite{DBLP:conf/nips/SocherCMN13}. NTN is the most expressive amongst the relational neural networks due to the high number of parameters and incredible modeling capability. The scoring function of NTN is defined as follows:

\begin{equation}
s(e_i,r_k,e_j) = u_{r_k}^{\top}f(e_i^{\top}W^{[1:s]}_{r_k}e_j + V_{r_k} [\vec{e_i};\vec{e_j}] + b_{r_k})
\end{equation}
where $f$ is a nonlinear function applied element-wise. $W^{[1:s]}_R \in \mathbb{R}^{n \times n \times k}$ is a tensor and each bilinear product $e_i^{\top}W^{[k]}_{r_k}e_j$ results in a scalar value. The scalar values across $s$ slices are concatenated to form the output vector $h \in \mathbb{R}^{k}$. The other parameters are in the standard form of neural networks.

\subsection{Embedding and Factorization Models}
It is worthy to note that there exist rich and extensive research that are concerned with latent embedding models for relational learning. These models are commonly known as \textit{translational embedding} models and are improvements over the highly influential work, TransE \cite{DBLP:conf/nips/BordesUGWY13}, proposed by Bordes et al. While there have been many extensions \cite{DBLP:conf/aaai/LinLSLZ15,DBLP:conf/aaai/WangZFC14,DBLP:conf/acl/JiHXL015,DBLP:conf/acl/0005HZ16,DBLP:conf/aaai/TayLH17} over this model, the key intuition behind translational embedding models is the concept of the translation principle, i.e., $|e_i + r_k - e_j|=0$ for a golden fact. 

On the other hand, the first and early latent embedding models for relational learning are mostly based on tensor factorization which exploits the fact that knowledge graphs inherently follow a 3D structure. Specifically, the general purpose Candecomp (CP) decomposition \cite{bro-parafac-1997} was first used in \cite{DBLP:conf/semweb/FranzSSS09} to rank triplets in a semantic web setting. Nickel et al. then proposed RESCAL \cite{DBLP:conf/www/NickelTK12,DBLP:conf/nips/JenattonRBO12} which is an improved tensor factorization technique based on the bilinear form for relational learning. Subsequently, extensions of RESCAL such as TRESCAL \cite{DBLP:conf/emnlp/ChangYYM14} and RSTE \cite{DBLP:conf/wsdm/TayLHB17} were proposed. 

It is easy to see that all latent embedding models can be interpreted as neural architectures. Specifically, in a review work by Nickel et al. \cite{DBLP:journals/pieee/Nickel0TG16}, the tensor factorization model RESCAL is casted as a neural network. As such, the scoring function of RESCAL is defined as:

\begin{equation}
g(e_i,r_k,e_j) = {W_k^{\top}}(\vec{e_j} \otimes \vec{e_i})
\end{equation}
where $W_k$ is a relation-specific parameter of the network and $\otimes$ is the tensor product. We can easily see that the parameter $W \in \mathbb{R}^{\abs{k} \times n \times n}$ is the middle tensor in the original RESCAL. Likewise, CP can also be casted similarly but we omit the discussion due to the lack of space. Similarly, translational embedding models, i.e., TransE, TransR, etc. are all neural architectures that optimize parameters via gradient descent.

\subsection{Multi-Task Learning (MTL)}
The key ide{}a of Multi-Task Learning (MTL) is to utilize the correlation between related tasks to improve the performance by learning the tasks in parallel. The success of multi-task learning \cite{caruana1998multitask} has spurred on many multi-task neural network models especially in the field of NLP \cite{DBLP:conf/ijcai/LiuQH16,DBLP:journals/corr/HashimotoXTS16,DBLP:conf/icml/CollobertW08}. While there are other variants of MTL such as multi-task feature learning \cite{DBLP:conf/ijcai/LiTLT15}, our work is concerned with shared representation learning, i.e., sharing parameters between two networks such that they benefit from training on multiple tasks. Many works \cite{DBLP:journals/corr/HashimotoXTS16,DBLP:conf/ijcai/LiuQH16} have shown that multi-task learning can lead to improved performance. This serves as the major motivation behind our work. To the best of our knowledge, we are the first multi-task learning approach in the context of knowledge graphs. 

\subsection{Multi-Source Relational Learning}
Many works have proposed incorporating an external source of information to enhance representation learning in KGs. Though similar, this is \textbf{different} from multi-task learning. In this case, external information is used to enhance learning but not as an extra task. As such, prediction of this additional information is usually not supported. For example, a recently fashionable line of research is concerned with joint representations of textual information with knowledge graphs \cite{DBLP:conf/emnlp/WangZFC14,DBLP:conf/emnlp/ZhongZWWC15,wang2016text}. There have been also many extensions of relational learning algorithms to various sources of external information such as hierarchical information \cite{DBLP:conf/ijcai/XieLS16}, logic rules \cite{DBLP:conf/emnlp/GuoWWWG16}, schema information \cite{DBLP:conf/emnlp/ChangYYM14}, path information \cite{DBLP:conf/emnlp/LinLLSRL15} and degree-aware property information \cite{DBLP:conf/sigir/JameelBS17}.

\subsection{Handling Attributes in Knowledge Graphs}
As mentioned, a prevalent problem is that none of the current models attempt to incorporate attribute information especially for non-discrete attributes.  In \cite{DBLP:conf/www/NickelTK12}, the authors proposed to use a separate matrix factorization to learn attributes. However, the authors only explore the possibility of adding tokenized text attributes. Furthermore, a huge problem is that each attribute would require an additional matrix factorization operation which can be impractical. A recent work, KR-EAR \cite{DBLP:conf/ijcai/LinLS16} is a translational embedding model that was proposed to model `attribute information'. Their main idea is that modeling attributes separately can result in better relational learning performance. However, their approach is concerned with relations that are actually attributes, e.g, \textit{gender} is an attribute that is often considered as a relation in most KGs. This means their approach only works with prior knowledge of knowing which relation should have been an attribute. Furthermore, their approach does not deal with non-discrete data types. As a result, though deceptively similar, our work is distinctly different from theirs and \textbf{cannot be meaningfully compared}. 

To the best of our knowledge, there is no approach that is able to elegantly incorporate continuous attribute information in the context of relational learning. This can be attributed to the inherent difficulty of dealing with non-discrete float data types in the context of the binary-natured KG. Essentially, attributes and their non-discrete values extend a KG to the fourth dimension.

\section{Multi-Task Knowledge Graph Neural Network (MT-KGNN)}
\begin{figure}[ht]
\begin{center}
\includegraphics[width=0.48\textwidth]{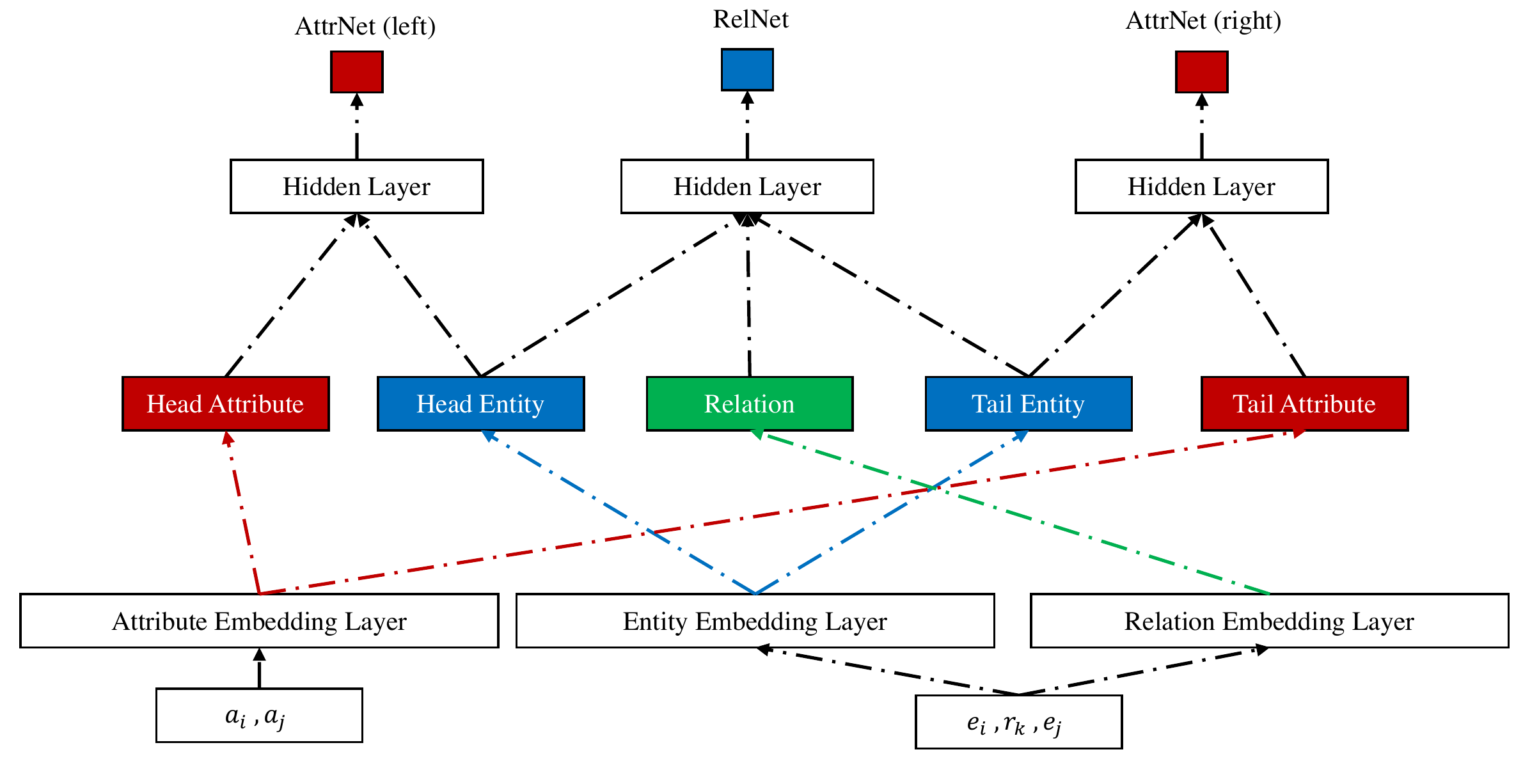}

\caption{Architecture of our proposed Multi-Task Knowledge Graph Neural Network (MT-KGNN).}
\label{Flowchart}
\end{center}
\end{figure}

In this section, we introduce MT-KGNN, our novel deep learning architecture for both relational learning and non-discrete attribute prediction on knowledge graphs. There are two networks in our model, namely the Relational Network (RelNet) and the Attribute Network (AttrNet). Figure \ref{Flowchart} describes the overall model architecture. Let us begin with an introduction of the new notation pertaining to the attribute data. 

\subsection{Relational Learning with Attributes}
Our work is concerned with utilizing attribute triplets to both improve the relational learning process and enable regression in a relational learning setting. First, we formally define the \textit{additional} notations that are supplementary to the earlier problem formulation. Let $A$ be the set of all non-discrete attributes such as height, price or population in a knowledge graph. The new notation for a knowledge graph can then be defined as $\Delta=(E,R,A)$. Note that attribute triplets are non-discrete, i.e., a non-discrete attribute triplet, $\psi_i \in \psi$ is defined as $(e_i, a_k, v)$ where $a_k$ is an attribute in $A$ and $v$ is a normalized continuous value from $[0,1]$. We assume that the range of each attribute can be sufficiently deduced from the training data. Any value at test time that exceeds the max-min of the normalization will automatically be casted to $0$ or $1$.

\subsection{Embedding Layer}

At training time, the inputs to the Relational Network are $[e_i,r_k,e_j,t]$ where $e_i, e_j \in \mathbb{R}^{n}$, $r_k \in \mathbb{R}^{m}$ and $t$, the target of classification is either $0$ or $1$. The inputs to the Attribute Network are $[a_i,v_i,a_j,v_j]$ where $a_i,a_j \in \mathbb{R}^{l}$ and $v_i,v_j \in [0,1]$. For simplicity, we will consider that $m=n=l$, i.e., all entity, relation and attribute embeddings have equal dimensions. The inputs of our model are discrete one-hot-encoded indices which will be passed into an embedding lookup layer to retrieve the corresponding vector representations. Here, $W_e \in \mathbb{R}^{\abs{E} \times n}$, $W_r \in \mathbb{R}^{\abs{R} \times n}$ and $W_a \in \mathbb{R}^{\abs{A} \times n}$ are the representations learned by the model. 

\subsection{Relational Network (RelNet)}
This section introduces the Relational Network (RelNet) used in our model that models the structural and relational aspect of the knowledge graph. RelNet is a simple concatenation of the triplet passed through a nonlinear transform and finally a linear transform with sigmoid activation. The RelNet component of our model is defined as follows:

\begin{equation}
g_{rel}(e_i,r_k,e_j) = \sigma(\vec{w}^{\top}f(\textbf{W}_d^{\top}[\vec{e_i};\vec{e_j};\vec{r_k}]) + b_{rel}) 
\end{equation}

\noindent where $f$ is the hyperbolic tangent function \textit{tanh}. $w \in \mathbb{R}^{h \times 1}$ and $\textbf{W}_d \in \mathbb{R}^{3n \times h}$ are parameters of the network. $\sigma$ is the sigmoid function and $b_{rel}$ is a scalar bias. 
 In order to train RelNet, we minimize the cross entropy loss function as follows:

\begin{equation}
L_{rel} = - \sum^{N}_{i=1} t_i \log g_{rel}(\xi_i) + (1-t_i) \log (1-g_{rel}(\xi_i))
\end{equation}
where $\xi_i$ denotes triplet $i$ in batch of size $N$. $t_i$ is a value of $\{0,1\}$. Cross entropy is a common loss function for non-continuous targets and is commonly used in many binary classification problems. At this point, it is good to note that RelNet remains identical to the ER-MLP model \cite{DBLP:conf/kdd/0001GHHLMSSZ14}. The contribution and novelty of our approach lies with the incorporation of the Attribute Network. 

\subsection{Attribute Network (AttrNet)}
This section introduces the Attribute Network (AttrNet) used in our model. Similar to RelNet, we train a single hidden layer network by concatenation of the attribute and entity embeddings to predict the continuous value which is a normalized value $\in [0,1]$. However, there are two entities in each relational triplet $\xi_i \in \xi$, namely the head and tail entities. Therefore, there are two \textit{sides} to AttrNet which are described as AttrNet (left) and AttrNet (right) as shown in Figure \ref{Flowchart}. The motivation of using two networks are as follows: since entities in knowledge graphs are generally considered as anti-symmetric relations, e.g., entities behave differently when they are at the head or tail position, we design the network in order to capture this relational information. As such, AttrNet optimizes a joint loss function of both regression tasks. We define the score functions as follows:

\begin{equation}
g_{h}(a_i) = \sigma(\vec{u}^{\top}f(B^{\top}[\vec{a_i};\vec{e_i}]) + b_{z_1}) 
\end{equation}

\begin{equation}
g_{t}(a_j) = \sigma(\vec{y}^{\top}f(C^{\top}[\vec{a_j};\vec{e_j}]) + b_{z_2}) 
\end{equation}
where $u,y \in \mathbb{R}^{h_a \times 1}$ and $B,C \in \mathbb{R}^{2n \times h_a}$ are parameters of AttrNet. $h_a$ is the size of the hidden layer and $b_{z_1},b_{z_2}$ are scalar biases. Similar to RelNet, the final output of each side of AttrNet is a scalar value. Moreover, since AttrNet models a continuous target and is formulated as a regression problem, we optimize the mean squared error (MSE) instead of the cross entropy loss function. The MSE loss function is given by:

\begin{equation}
MSE(s,s^{*}) = \frac{1}{N} \sum^{N}_{i=1} (s_i - s^*_i)^2
\end{equation}
Let $L_{head}$ and $L_{tail}$ be the loss functions for each side of the Attribute Network which both use the MSE loss function. The overall loss of the Attribute Network is simply the sum of both sides of the network. 
\begin{equation}
L_{attr} = L_{head} + L_{tail}
\end{equation}
\begin{equation}
L_{head} = MSE(g_{h}(a_i), (a_i)^{*})
\end{equation}
\begin{equation}
L_{tail} = MSE(g_{t}(a_j), (a_j)^{*})
\end{equation}
where $(a_i)^{*}, (a_j)^{*}$ are the ground truth labels. Similar to RelNet, we apply the constraints $\abs{e}_2 \leq 1$ and $\abs{a}_2 \leq 1$ for regularization. 

\begin{table*}[ht]
  \centering
\small
    \begin{tabular}{c|c|c|c}
   \hline
    Model & \multicolumn{1}{l|}{Space Complexity} & \multicolumn{1}{l|}{\# Params} & \multicolumn{1}{l}{Operations (Time Complexity)}\\
    \hline
    RESCAL \cite{DBLP:conf/www/NickelTK12,DBLP:conf/icml/NickelTK11}&   $N_e m + N_r n^2$ &  1.278M & $(m^2 + m) N_t$\\
    TransE \cite{DBLP:conf/nips/BordesUGWY13} &   $N_e m + N_r n$    &  1.202M & $N_t$ \\
    TransR \cite{DBLP:conf/aaai/LinLSLZ15} &    $N_e m + N_r (mn)$   & 1.278M & $2mn N_t$\\
    ER-MLP \cite{DBLP:conf/kdd/0001GHHLMSSZ14} & $N_e m + N_r n  + ((2m+n) \times h) + h$ & 1.217M & $(3mh + h) N_t$\\
    NTN  \cite{DBLP:conf/nips/SocherCMN13} &   $N_e m + N_r (n^2 s + 2ns + 2s)$   & 1.523M & 
    $((m^2 + m) \: s + 2ms + s) \: N_t$\\
    \hline
    RelNet  & $N_e m + N_r n + N_a n^{*}  + ((2m+n) \times h) + h $ & 1.278M & $((2m+n) \: h + h) N_t$\\
    AttrNet &$N_e m + N_a n^{*} +  2((m+n^{*}) \times h) + h$  & 1.221M & $(2 + k) (m + n^{*}) N_{t} $\\
    MT-KGNN & $N_e m + N_r n + N_a n^{*}  + ((2m+n) \times h) + 2h + 2((m+n^{*}) \times h)$     & 1.237M &  $((2 + k) (m + n^{*}) + ((2m+n) \: h + h) )N_{t} $\\
    \hline
    \end{tabular}%
  
    \caption{Space complexity analysis of our proposed approach relative to other state-of-the-art relational learning models. $N_e, N_r, N_a$ are the number of entities, relations and attributes in $\Delta$. $N_t$ is the number of relational triplets. $m$ is the dimensionality of entity embeddings while $n$ is the dimensions of the relation embeddings. $n^{*}$ is the dimensionality of the attribute embeddings. $h$ is the dimensionality of the hidden layer in ER-MLP and our proposed approach. $k$ is the hyperparameter for AST. $s$ is the number of tensor slices in the NTN model. }
    \label{tab:complexity}%
\end{table*}%

\subsection{Multi-Task Optimization and Learning}

In this section, we introduce a multi-task learning scheme for training both AttrNet and RelNet together. Our proposed \textit{Multi-Task Knowledge Graph Neural Network} (MT-KGNN) is a multi-task learning scheme for simultaneously learning from relational and attribute information. Algorithm \ref{mtrrn_algorithm} details the multi-task learning process of our approach.  

\subsubsection{Multi-Task Learning Scheme}
Similar to many other multi-task learning techniques \cite{DBLP:conf/ijcai/LiuQH16}, both RelNet and AttrNet are trained in an alternating fashion. Note that the input vector of AttrNet contains attributes belonging to the head and tail entities contained within the relational triplet. This not only optimizes the parameters of the network with both relational and attribute information but also allows AttrNet to gain some structural and relational context. The overall learning process can be summarized as follows:
\begin{itemize}
\item \textbf{Relational Training (RT)} - Sample a relational triplet and train RelNet (line 7).
\item \textbf{Attribute Training (AT)} - Build input for AttrNet, i.e., for each entity, head and tail in the triplet, randomly sample an attribute to form the input vector. If either entity has no attribute, then append a zero vector, i.e., [0,0] to the side of AttrNet that corresponds to this entity (line 9). The parameters of AttrNet are then updated as we train AttrNet (line 10).
\end{itemize}

\begin{algorithm}[H]
\small
\caption{Multi-Task Knowledge Graph Neural Network}
\label{mtrrn_algorithm}
\begin{algorithmic}[1]

\Require{KG $\Delta$, rel triplets $\xi$, attr triplets $\psi$, batch size=$\beta$, model hyperparameters}
\Ensure{Trained Model Parameters $\theta=\{\theta_r, \theta_a\}$}
  \State $E,R,A \gets$ all entities, relations and attributes in $\xi$ respectively 
  \State Initialize uniform $(-\frac{6}{\sqrt{k}},\frac{6}{\sqrt{k}})$ for $e_i \in E, r_i \in R, a_i \in A$ 
  \State Initialize random normal N(0,1) for $\theta_{r} = \{\vec{w},\vec{v},D\}$
  \State Initialize random normal N(0,1) for $\theta_{a} = \{\vec{u},\vec{v},B,C\}$
  \For{each epoch}
    \State $\chi_r \gets$ shuffle($\chi_r$) 
    \State $\chi_r \gets $ sample batch of size $\beta$ from $\xi$
    \State $\theta_r,E,R \gets$ Update w.r.t $L_{rel}(\chi_r)$  
    \State $\chi_a \gets$ buildAttributes($\psi,\chi_r$)
    \State $\theta_a,E,A \gets$ Update w.r.t $L_{attr}(\chi_a)$  
    
    \For{each $k$}
      \State $i \gets$ sample random attribute from $A$
      \State $\psi^{i} \gets$ select all attr triplets containing $i$
      \State $\chi_a^{i} \gets$ sample batch of size $\beta$ from $\psi^{i}$
      \State $\theta_a,E,A \gets$ Update w.r.t $L_{attr}(\chi_a^i)$ 
    \EndFor

\EndFor

\end{algorithmic}
\end{algorithm}

\subsubsection{Attribute Specific Training (AST)}
 Next, we introduce an alternative method for training attributes which corresponds to line 11 to line 15 in Algorithm \ref{mtrrn_algorithm}. We call this process \textit{Attribute Specific Training} (AST) which may be used \textbf{in conjunction} with Attribute Training (AT) or to \textbf{replace} the Attribute Training step mentioned earlier. In this step, we train AttrNet an additional $k$ times in an \textbf{attribute specific} manner, i.e., for $k$ times, we sample an attribute and select $\beta$ attribute triplets, and run them through AttrNet (both left and right). The key intuition here is that we want to supply the network with attribute information in an \textit{orderly} attribute-specific and focused fashion which allows the model to concentrate on a single attribute. Furthermore, this also controls the amount of attribute information that is learned relative to the relational information since each AST call essentially trains the network once with attribute information. Empirically, we found that for both triplet classification and attribute value prediction, using AST improves the performance.

\subsubsection{Regularization}
Finally, the last \textit{post-processing} step that our model takes is to normalize all embeddings to be constrained within the Euclidean ball. At the end of training RelNet and AttrNet (multiple times due to AST), we apply the constraints of $\abs{e_{*}}_2 \leq 1$, $\abs{r_{*}}_2 \leq 1$ and $\abs{a_{*}}_2 \leq 1$ for regularization and preventing overfitting. Moreover, we apply dropout of $d=0.5$ at the hidden layers of both RelNet and AttrNet.

\subsubsection{Shared Representations}
In a nutshell, the core motivation to this multi-task scheme is as follows: Since the entity embeddings $\vec{e_i} \in E$ are used in both $L_{attr}$ and $L_{rel}$ when RelNet and AttrNet are trained separately, the entity embeddings are updated from both tasks. As such, the entity embeddings are learned from both the relational task as well as the attribute task.

\subsection{Complexity Analysis}

In this section, we study the complexity of our proposed model relative to many other state-of-the-art models. Table \ref{tab:complexity} reports the space and time complexity of all models. Additionally, we include an estimate on the number of parameters on the YAGO subgraph that we use in our experiments. Our proposed MT-KGNN does not incur much parameter cost over the baseline ER-MLP. MT-KGNN also shares entity embeddings across RelNet and AttrNet. Therefore, the additional cost is only derived from attribute embeddings (which are usually much smaller) and additional hidden layer parameters for AttrNet. 

Pertaining to time complexity, our model is equivalent to training ER-MLP multiple times (which is dependent on the hyperparameter $k$ if AST is used). Similar to space complexity, this is still less operations compared to models requiring operations in quadratic scale such as TransR, NTN or RESCAL. 


\section{Experimental Evaluation}
We conduct two experiments to evaluate our model. First, we provide an experimental analysis of our model in the standard relational learning benchmark task of Relational Triplet Classification. This is in concert with many works such as \cite{DBLP:conf/nips/SocherCMN13,DBLP:conf/aaai/LinLSLZ15}. Second, we design another experiment of attribute value prediction to investigate our model's ability to perform regression-like prediction which contributes to the key feature of our model.

\subsection{Datasets}
We use two datasets in our experiments constructed from two carefully selected web-scale KGs.
\begin{itemize}
\item \textbf{YAGO} \cite{DBLP:conf/ijcai/HoffartSBW13} is a semantic knowledge base that aggregates data from various sources including WikiPedia. YAGO contains many famous places and people which naturally contain many attribute values such as height, weight, age, population size, etc. We construct a subset of YAGO by removing entities that appear less than $25$ times and further filtered away entities that do not contain attribute information. We name our dataset YG24K since it contains $\approx 24$K entities.

\item \textbf{Freebase} \cite{DBLP:conf/aaai/BollackerCT07} is a widely used benchmark knowledge base for many relational learning tasks. We use the dataset dump \textit{Easy Freebase} \cite{DBLP:conf/www/BastBBH14} since the public API is no longer available. Due to the large size of Freebase, we extract a domain-specific dataset from Easy Freebase involving organizations. The construction of this dataset is detailed as follows:

\begin{itemize}
\item We randomly selected entities of type \textit{organization} and selected $22$ relations that are closely related to the domain such as (industry), (school\_type), (academic\_adviser)
and (founded\_by).

 \item Subsequently, using these seed entity nodes we randomly included relational and attribute triplets within a \textit{k-hop} distance. Examples of attributes in this dataset include (num\_employees), (num\_postgraduates) and (height), etc.  The end result is a dataset with $28$K entities which we call FB28K. Note that since we applied the random \textit{k-hop} selection of nodes in the knowledge graph, the extracted subgraph would contain entities and relations across several domains. 
\end{itemize}
\end{itemize}

\noindent For all datasets, we split the relational and attribute triplets into an 80/10/10 split of train/development/test. Subsequently, we filter triplets in the test and development sets that contain entities not found in the training set. The statistics of the two datasets is given in Table \ref{tab:dataset}.

 \begin{table}[htbp]
   \centering
   
     \begin{tabular}{c|ccccc}
     \hline
           & \# Trip & \# Attr Trip & \# Ent & \# Rel & \# Attr \\
           \hline
     YG24K &  142,701     &   33,550    &   24,831    & 31      & 18 \\
     FB28K & 160,850 & 71,776 & 28,622 & 22      & 77 \\

     \hline
     \end{tabular}%
      \caption{Summary of dataset characteristics.}
   \label{tab:dataset}%
 \end{table}%
On FB28K, $33\%$ of the total number of triplets are attribute triplets. On the other hand, $15\%$ of the total number of triplets in YG24K are attribute triplets. 

\subsection{Algorithms Compared}
For our experiments, we compare our proposed MT-KGNN approach with many state-of-the-art models. 
\begin{itemize}
\item \textbf{CANDECOMP (CP)} \cite{DBLP:conf/semweb/FranzSSS09} is a classical tensor modeling technique that scores each triplet using $s(e_i, r_k, e_j) = e_i \: \odot r_k \: \odot e_j$.
\item \textbf{RESCAL} \cite{DBLP:conf/icml/NickelTK11} is a tensor factorization approach based on bilinear tensor products. The scoring function of RESCAL is ${W_k^{\top}}(\vec{e_j} \otimes \vec{e_i})$.
\item \textbf{TransE} \cite{DBLP:conf/nips/BordesUGWY13} is a translational embedding model. A seminal work by Bordes et al. that models relational triplets with $|h+r-t| \approx 0$.
\item \textbf{TransR} \cite{DBLP:conf/aaai/LinLSLZ15} is an extension of TransE that proposes using matrix projection to model relation specific vector space. The scoring function of TransR is $|M_rh+r-M_{t}t| \approx 0$.
\item \textbf{ER-MLP} \cite{DBLP:conf/kdd/0001GHHLMSSZ14} is the baseline neural network approach that learns representations via a concatenation operator of triplets. ER-MLP enables a direct comparison with our model to study the effects of incorporating non-discrete attributes. 
\item \textbf{Neural Tensor Network (NTN)} \cite{DBLP:conf/nips/SocherCMN13} is a highly expressive model that combines a bilinear tensor product with a MLP. We use a setting of $s=4$ following \cite{DBLP:conf/nips/SocherCMN13} where $s$ is the number of tensor slices. 
\item \textbf{Multi-task Knowledge Graph Neural Network (MT-KGNN)} is the proposed approach in this paper. We use $k=4$ for AST (Attribute Specific Training) in conjunction with AT (Attribute Training).  
\end{itemize}

\subsection{Implementation Details}
We implement all models ourselves in TensorFlow \cite{tensorflow2015-whitepaper}. All algorithms are optimized with the Adam optimizer \cite{kingma2014adam} with an initial learning rate of $10^{-3}$ and have their embeddings (entity, relation and attribute when applicable) set to $50$ dimensions. All algorithms optimize the sigmoid cross entropy loss function except TransE and TransR which we minimize the pairwise hinge loss as stated in their original papers. For both RESCAL and CP, we consider the neural network adaptation \cite{DBLP:journals/pieee/Nickel0TG16}. For ER-MLP and MT-KGNN, we set the size of \textit{all} hidden layers to be $100$, and dropout to be $d=0.5$ with the $tanh$ activation function. All models have their entity and relation embeddings constrained to $\abs{e}_{2} \leq 1$ and $\abs{r}_{2} \leq 1$. If relation embeddings are matrices instead of vectors, the norm is restricted to $3$ instead. The batch size is set to $500$ and the number of iterations is fixed to $500$ iterations for all models. Additionally, due to the sensitivity to hyperparameters of TransE and TransR, we consider margins amongst $\{1.0, 2.0, 4.0\}$ and take the model with the best performance on the development set. 
All experiments were conducted on a machine running Linux with a NVIDIA GTX1070 GPU. 

\subsection{Experiment 1 - Relational Triplet Classification}
In this experiment, we demonstrate the ability of our model on a standard relational learning benchmark of triplet classification which is essentially a binary classification task. This benchmark task has been widely adopted in many works \cite{DBLP:conf/nips/SocherCMN13,DBLP:conf/kdd/0001GHHLMSSZ14,DBLP:conf/acl/JiHXL015}. The task is as follows: Given a triplet of $(h,r,t)$, we classify it as being $1$ or $0$ (true or false). We follow the experimental procedure of \cite{DBLP:conf/nips/SocherCMN13} closely. The aim of this experiment is to show that additional attribute information help in standard relational learning tasks.

\subsubsection{Evaluation Procedure and Metrics}
In order to perform classification, we perform negative sampling following \cite{DBLP:conf/nips/SocherCMN13}. For each positive triplet in our train/development/test sets, we include a corrupt triplet. We do this by randomly replacing either the head or tail entity. The final positive to negative ratio is therefore 1:1. For evaluation purposes, we use two popular metrics for binary classification, namely accuracy and AUC (Area Under Curve). Regarding the metric of accuracy, we determine a threshold where scores above are classified as positive and vice versa. This threshold is determined by maximizing the accuracy across the development set.

\subsubsection{Experimental Results}
Table \ref{tab:tripletexp} shows the results of the triplet classification experiment. Our proposed MT-KGNN achieves the state-of-the-art performance on both YG24K and FB28K. Our proposed approach outperforms the baseline ER-MLP by at least $2\%-3\%$ and models like TransE by $3\%-4\%$. This suggests that our multi-task learning scheme is highly effective for the relational learning task. Evidently, the attribute information is able to increase the accuracy by $3\%$ in both datasets and as a whole achieve about an improvement of $>5\%-6\%$ accuracy as compared to models such as CP or TransR. Pertaining to the relative performance of the compared baselines, the best performing result is obtained from ER-MLP. On the other hand, NTN produces poorer results. We also observe that models of higher complexity seem to perform worse than their counterparts with smaller number of parameters. For example, RESCAL performs better than NTN, and TransE outperforms TransR.

 \begin{table}[htbp]
   \centering
     \begin{tabular}{c|cc|cc}
     \hline
           & \multicolumn{2}{c|}{YG24K} & \multicolumn{2}{c}{FB28K}  \\
           \hline
           & Accuracy & AUC   & Accuracy & AUC     \\
           \hline
     CP    & 83.6  & 91.6  &       83.7 &  89.2            \\
     RESCAL & 87.6  & 94.1     &      84.6 &  90.8       \\
     TransE & 86.1  & 93.2 &       85.8 & 91.7          \\
     TransR &  84.8 & 89.3      &       84.6 & 88.7         \\
     ER-MLP  & \underline{88.4} &  \underline{94.1}  &      \underline{86.2}  & \underline{91.3}         \\
     NTN   &  85.1  & 91.8          &      85.2 & 90.0              \\
     \hline

     MT-KGNN & \textbf{91.0}  & \textbf{96.2} &       \textbf{89.6} &  \textbf{94.2}           \\
     \hline
     \end{tabular}%
     \caption{Experimental results on relational triplet classification. Our proposed MT-KGNN achieves the state-of-the-art
     performance. Best result is in boldface and second best is underlined. }
   \label{tab:tripletexp}%
 \end{table}%

\subsection{Experiment 2 - Attribute Value Prediction}
In this task, we evaluate the capability of our model to perform regression, i.e., predictions of non-discrete attributes in KGs. For all compared baselines other than ours, prediction is performed using attribute-specific \textit{Linear Regression} classifiers trained using the learned embeddings as features. Apart from validating the performance of our proposed model, the \textit{additional} aims of this experiment are as follows: 
\begin{enumerate}
\item \textbf{RQ1}: In the context of attribute prediction, are there useful information encoded in entity embeddings produced by relational learning models?
\item \textbf{RQ2:} Is there any difference between different relational learning models in encoding attribute information, e.g., is TransE or RESCAL better than ER-MLP when the learned embeddings are used as features? 
\item \textbf{RQ3:} Which is better for encoding attribute information? Knowledge graph embeddings or word embeddings?
\end{enumerate}
The impact of this experiment is meaningful as it derives insight regarding the plausibility of knowledge graph embeddings as features for standard machine learning tasks. Note that the usage of supervised classifiers for evaluating the quality of embeddings has been substantiated in many works \cite{DBLP:conf/emnlp/GuptaBBP15,DBLP:conf/emnlp/LuuTHN16}.

\subsubsection{Implementation Details}
For all KG embedding methods, we train embeddings of size $n=50$ for $500$ iterations with the same settings as Experiment 1. By using these embeddings as features, we train a separate linear regression classifier using Stochastic Gradient Descent (SGD) with learning rate amongst $\{10^{-2}, 10^{-3},10^{-4} \}$ for $25$ epochs since we empirically found that $25$ epochs are sufficient for the model to converge on the development set. Similar to the first experiment, we report the performance of the model that performed best on the development set. 

\subsubsection{Additional Baselines}
Aside from knowledge graph embeddings, we also introduce several baselines.
\begin{itemize}
\item \textbf{Random Guess (R-GUESS)} randomly generates a value $v \in [0,1]$ as a prediction.  
\item \textbf{Random Init (R-INIT)} trains the supervised classifier on an embedding matrix that is initialized randomly with $U(-0.01,0.01)$. As such, the purpose is to detect if the knowledge graph embedding is producing better performance than random. 
\item \textbf{SkipGram} \cite{DBLP:conf/nips/MikolovSCCD13} or also known as Word2Vec is a highly popular language modeling approach that learns continuous vector representations of words from textual documents. For SkipGram, we use the pre-trained vectors of $300$ dimensions released by Google. For each entity in the KG, we form a term embedding by averaging the vectors of multiple words. 

\end{itemize}
For our proposed approach, we use the MT-KGNN network directly for prediction. For this task, we removed the optional attribute training step and simply use AST with $k=4$. 

\subsubsection{Evaluation Metrics}
In this task, we use three popular evaluation metrics for evaluating the regression problem, namely the RMSE (Root Mean Square Error), Mean Absolute Error (MAE) and $R^{2}$ (R squared). These metrics are defined as follows:
\begin{equation}
RMSE = \sqrt{\frac{1}{n} \sum^{n}_{i=1} (y_i - \hat{y_i})^2}
\end{equation}
  
\begin{equation}
 MAE = \frac{1}{n} \sum^{n}_{i=1} \: \abs{y_i -\hat{y_i}}
\end{equation}
\begin{equation}
R^{2} = 1 - \frac{\sum^{n}_{i=1} (y_i-\hat{y_i})^{2}}{\sum^{n}_{i=1} {(\hat{y_i}-\bar{y_i}})^{2}}
\end{equation}
where $y_i$ is the ground truth, $\hat{y_i}$ is the prediction and $\bar{y_i}$ is the mean $y$ value. RMSE and MAE measure the fit of the prediction by reporting the amount of \textit{error} relative to the ground truth. The main difference is that RMSE penalizes larger mistakes to a larger extent due to its quadratic nature. The $R^2$ metric is a common metric in linear regression that expresses the percentage of the variation in the ground truth that is explained by the prediction. Note that the $R ^{2}$ metric can result in negative values which essentially means that the prediction fits worse than a horizontal line. 

\subsubsection{Experimental Results}

\newcolumntype{C}[1]{>{\centering}m{#1}}

\begin{table}[htbp]
  \centering
  
    \begin{tabular}{c|ccc|ccc}
    \hline
          & \multicolumn{3}{c|}{\textbf{YG24K}} & \multicolumn{3}{c}{\textbf{FB28K}} \\
          \hline
          & RMSE  & MAE   & $R^2$   & RMSE  & MAE   & $R^2$ \\
          \hline
    R-GUESS & 0.374 & 0.308 & -0.777 & 0.371 & 0.306 & -0.650 \\
    R-INIT & 0.444 & 0.363 & -1.322 & 0.416 & 0.339 & -1.050 \\
    SkipGram & 0.302 & 0.257 & -0.145 & 0.295 & 0.254 & -0.040 \\
    \hline
    CP    & 0.289 & 0.251 & -0.004 & 0.291 & 0.253 & \underline{0.003} \\
    TransE & \underline{0.287} & \underline{0.248} & -0.004 & 0.289 & 0.251 & 0.002 \\
    TransR & 0.293 & 0.256 & 0.002 & 0.287 & 0.248 & -0.003 \\
    RESCAL & 0.290 & 0.251 & -0.006 & 0.292 & 0.253 & -0.004 \\
    ER-MLP & 0.288 & 0.249 & 0.000 & 0.289 & 0.250 & \underline{0.003} \\
    NTN   & 0.289 & \underline{0.248} & \underline{0.002} & \underline{0.286} & \underline{0.247} & -0.002 \\
    \hline
    MT-KGNN & \textbf{0.065}	& \textbf{0.013} &	\textbf{0.879} & \textbf{0.105} &	\textbf{0.052}	& \textbf{0.750} \\
    \hline
    \end{tabular}%
    \caption{Experimental results for attribute value prediction. Best result is in boldface and second best is underlined. Our proposed MT-KGNN achieves significantly better performance.}
  \label{tab:attr_results}%
\end{table}%

Table \ref{tab:attr_results} reports the results of the attribute value prediction. Firstly, we observe that MT-KGNN achieves the best results. Moreover, there is a distinct and significant margin as compared to all other relational learning methods. Notably, there is about $\approx 30\%$ decrease in RMSE across both datasets relative to almost all relational learning embedding models. In addition, our model is the only model that achieves a decent $R^2$ score ($>0$). Hence, to the best of our knowledge, our approach is the only model that is able to predict attribute values in knowledge graphs by exploiting relational information without the use of any feature engineering. Moreover, aside from relation-specific linear regression models that have to be built for \textbf{each} relation, our model handles this with only a single model that works with all attributes collectively. 

Secondly, we observe that relational learning models perform significantly better over R-INIT and R-GUESS. As such, we can conclude that there is at least some useful information encoded in relational learning models that are useful for attribute prediction. This answers \textbf{RQ1} though we believe that the information encoded to be minimal. Thirdly, we observe that the performance of all relational learning baselines are relatively indistinguishable from one another. The performance of TransE, TransR, NTN, RESCAL, ER-MLP and CP produce approximately similar results. As such, pertaining to \textbf{RQ2}, we conclude that the performance of different relational learning models do not differ much from each other. Finally, the performance of SkipGram is slightly worse than relational learning models. To answer \textbf{RQ3}, we are able to conclude, based on the empirical evidence, that relational learning models produce slightly more useful features compared to word embeddings.

\subsection{Discussion and Analysis}
The aim of this section is to provide additional insights pertaining to our proposed approach. Firstly, we perform an ablation study to show the relative effect of RelNet and AST. Secondly, we extract the attribute embeddings from MT-KGNN and analyze them. 
\subsubsection{Ablation Studies}

Table \ref{tab:abstudy} reports our ablation study on FB28K for attribute value prediction. Specifically, we removed RelNet and AST from MT-KGNN to see how much each process contributes to the overall performance of MT-KGNN. Naturally, AST would be expected to be more important. As such, the main investigation is pertaining to the effect of relational triplets on attribute prediction. We observe that there are indeed observable improvements in performance when using RelNet. For instance, the $R^{2}$ metric is improved by $17$ points while MAE and RMSE are also improved considerably. 
\begin{table}[htbp]
  \centering

    \begin{tabular}{l|lll}
    \hline
          & RMSE  & MAE   & $R^2$ \\
          \hline
            MT-KGNN & 0.105 & 0.0518 & 0.750 \\
    -\textit{AST} & 0.241 (+0.14) & 0.116 (+0.06) & -0.300 (-1.05) \\
    -\textit{RelNet} & 0.137 (+0.03) & 0.100 (+0.05) & 0.578 (-0.17) \\
    \hline
    \end{tabular}%

    \caption{Ablation study on FB28K. Results show that removing RelNet degrades the performance.}
     \label{tab:abstudy}%
\end{table}

\vspace{-2.5em}

\subsubsection{Qualitative Analysis}

 \begin{table}[htbp]
   \centering
   
     \begin{tabular}{l|l}
     \hline
     \multicolumn{1}{l|}{Attribute} & \multicolumn{1}{l}{Nearest Neighbors } \\
     \hline
          num$\_$postgraduates & num$\_$undergraduates, num$\_$employees \\
          career$\_$losses$\_$doubles & career$\_$losses$\_$singles, num$\_$tennis$\_$titles \\
           household$\_$count & num$\_$mortgages, num$\_$foreclosures \\
          floors & retail$\_$store$\_$space, structural$\_$height  \\
          length & episode$\_$number, season$\_$number \\
          \hline
     \end{tabular}%
     \caption{Nearest neighbors of attribute embeddings in FB28K.}
   \label{tab:attrembed}%
 \end{table}%

In this section, we inspect the learned attribute representations of our proposed MT-KGNN. Using cosine similarity, we find the nearest neighbors of the attribute embeddings. Table \ref{tab:attrembed} shows some examples. It can be clearly seen semantically relevant attributes are closer to each other in the vector space. For example, \textit{num\_postgraduates} and \textit{num\_undergraduates} are close in the vector space probably due to their belonging to similar entities (universities). Additionally, attributes such as \textit{floors}, \textit{space} and \textit{height} are attributes belonging to building entities. As such, this ascertains the representation learning ability of our model since semantically similar attributes are clustered together in the vector space.

\section{Conclusion}
We introduced a novel concept of incorporating non-discrete attribute values in relational learning. Non-discrete attributes have traditionally been challenging to deal with as they do not fit intuitively into the binary-nature of KGs. As such, our proposed MT-KGNN is a multi-task neural architecture that can elegantly incorporate and leverage this information. It has demonstrated the state-of-the-art performance in the relational task of triplet classification and attribute value prediction. In both tasks, we observe that the relational and attribute information are complementary to each other.

\bibliographystyle{ACM-Reference-Format}
\bibliography{references}

\end{document}